\documentclass[letterpaper]{article} 
\usepackage{aaai2027}  
\usepackage[hyphens]{url}  
\usepackage{graphicx} 
\usepackage{natbib}  
\usepackage{caption} 
\usepackage{algorithm}
\usepackage{subcaption}
\usepackage{multirow}
\usepackage{latexsym}
 \usepackage{amsmath}
\usepackage{aaai2027} 
\usepackage[hyphens]{url} 
\usepackage{graphicx} 
\usepackage{natbib}  
\usepackage{caption}  
\usepackage[T1]{fontenc}
\usepackage{algpseudocodex}
\usepackage{newfloat}
\usepackage{listings}
\DeclareCaptionStyle{ruled}{labelfont=normalfont,labelsep=colon,strut=off} 
\floatstyle{ruled}
\newfloat{listing}{tb}{lst}{}
\floatname{listing}{Listing}

\usepackage{booktabs}
\usepackage{tcolorbox}
\tcbuselibrary{breakable}
\title{Long-form Video-to-Text Summarization using Recursive Language Models}
\author{
    Alessia Intini\textsuperscript{\rm 1}, Lorenzo Vaiani\textsuperscript{\rm 1}, Enrico Chen\textsuperscript{\rm 1}, Luca Cagliero\textsuperscript{\rm 1}
}
\affiliations{
    \textsuperscript{\rm 1}Politecnico di Torino\\

    \{name.surname\}@polito.it
}

\begin{document}

\maketitle

\begin{abstract}
While VideoLLMs have made significant progress in understanding short video benchmarks, they often struggle to summarize long-form ones. This is mainly due to performance degradation caused by long-context prompts, the inherent difficulty of preserving temporal continuity and a lack of semantically rich concepts in event descriptions. 

To tackle these issues, this paper introduces the first recursive Video Language Summarization Model (RVLSM) designed for video processing and tailored to long-form video-to-text summarization. RVLSM recursively decomposes videos into semantically coherent segments and delegates chunk-level reasoning to VideoLLM sub-models and external tools. 
It employs (1) Semantic Temporal Decomposition (STD), which preserves temporal coherence and continuity
through a scene-aware chunking strategy that dynamically partitions videos according to semantic boundaries, and 
(2) Adaptive Multimodal Tool Calling (AMTC), which ensures semantic richness of the summary via a routing mechanism that selectively invokes auxiliary modules such as action recognition, optical character recognition, automatic speech recognition, and object detection. 

Experiments on the VideoXum and MixTyVSum benchmarks show that RVLSM consistently improves over zero-shot VideoLLM baselines across multiple model settings and performance metrics, and is competitive with a state-of-the-art fine-tuned model for video-to-text summarization. Importantly, the named entities in the RVLSM's summaries align more closely with those detected in the ground-truth summaries than those detected by baseline and state-of-the-art methods, confirming the saliency of the event descriptions it generates. 


\end{abstract}

\section{Introduction}

\begin{figure}[h!]
    \centering
    \includegraphics[width=0.30\textwidth]{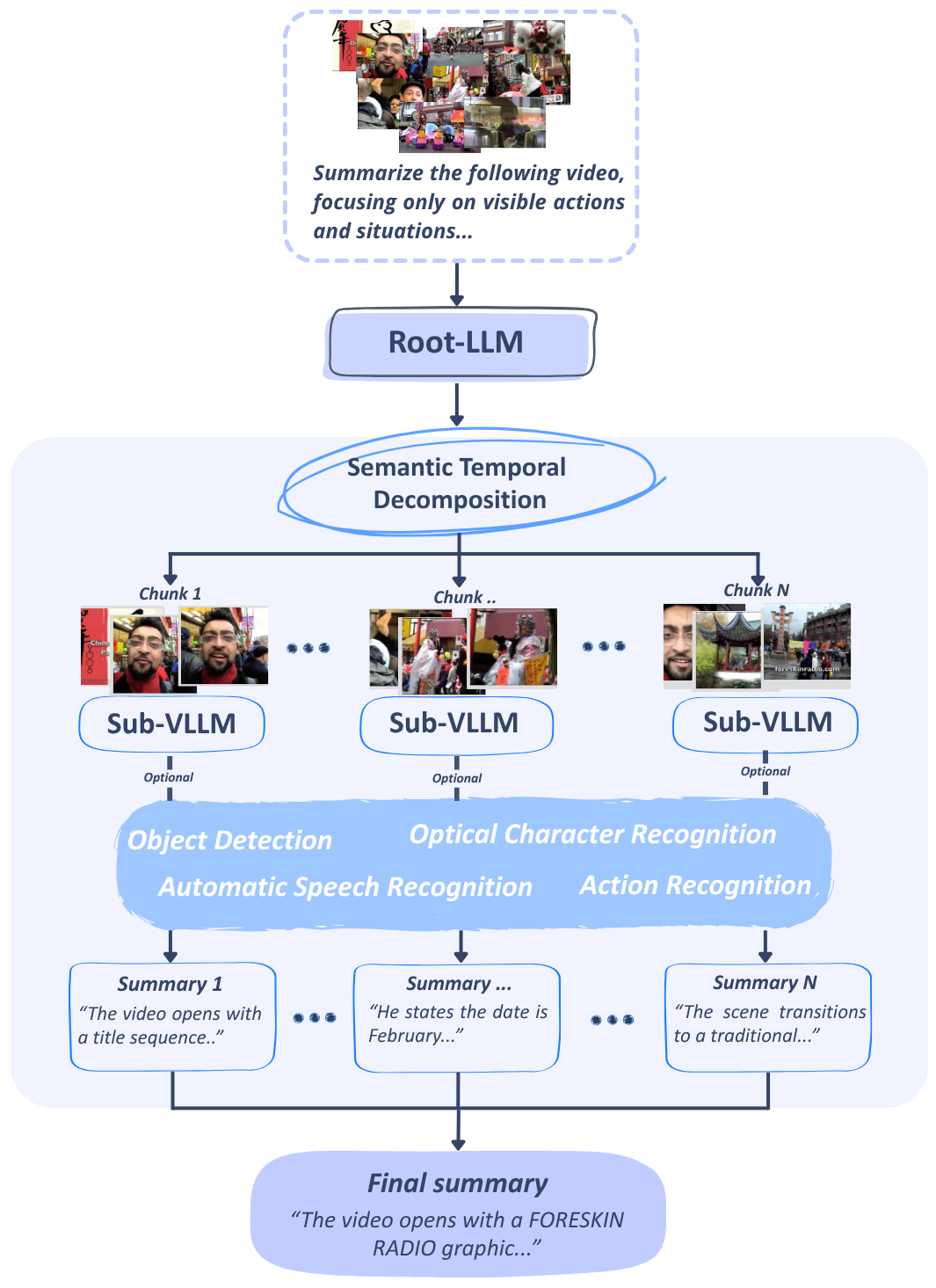}
\caption{Sketch of the RVLSM paradigm. The Root-LLM decomposes the long video, recursively invokes Sub-LLMs and external tools (action recognition, object detection, etc.), and composes the summary from the intermediate outcomes.}
    \label{fig:framework}
\end{figure}

Recent advances in VideoLLMs have significantly improved their understanding capabilities across different tasks, including video-to-text summarization. To create meaningful summaries of long-form videos,
models must identify and correlate key events and behaviors that occur over a
long period of time, which requires effectively capturing and
processing long-range temporal dependencies~\cite{11113316}.

Despite recent improvements, the performance of VideoLLMs on long-form videos is challenged by the following issues:

\begin{itemize}
    \item \textbf{Information compression}: long videos are frequently compressed through sparse frame sampling, temporal pooling, or fixed-duration chunking, forcing models to reason over aggressively condensed visual representations~\cite{Wang2025ICCV}. While often sufficient for short clips, these strategies become increasingly problematic for long-horizon and dense-event videos.
    \item \textbf{Context saturation}: LLMs are prone to hallucination and performance drops when their context window is largely filled~\cite{NEURIPS2024_6ffdbbe3}. Since the position of multimodal information also matters, this is typically exacerbated by the \emph{lost-in-the-middle} issue.
    \item \textbf{Temporal coherence degradation}: on videos spanning multiple scenes, subjects, or actions, VideoLLMs struggle to preserve the temporal continuity of the event structure and the semantic saliency of event descriptions. Long-form video-to-text summarization thus requires reasoning over temporally distributed events while maintaining chronological consistency and exhaustive coverage of the underlying concepts~\cite{Niu_2025_CVPR}.
    \item \textbf{Limited semantic richness of the event descriptions}: summaries typically mention the key events occurring in the video, but the semantic concepts in the textual descriptions are not necessarily aligned with human expectations~\cite{cagliero-etal-2025-detecting}.
\end{itemize}

To tackle the above issues, in this paper we present the first adaptation of the Recursive Language Models (RLMs)~\cite{zhang2026recursivelanguagemodels} towards long-form video-to-text summarization. 
RLM is a recently proposed inference paradigm for textual LLMs, 
where long textual context is treated as an external object inside persistent Read-Eval-Print Loop (REPL) environments, allowing a root-LLM to 
recursively invoke sub-VLLMs or external tools over portions of the input through programmatic decomposition.
Although they have proven to be effective in document summarization~\cite{zhang2026recursivelanguagemodels}, 
the existing framework is limited to the textual modality, overlooking the unique challenges of video-to-text summarization.

We present \textit{Recursive Video Language Summarization Models} (\textit{RVLSMs}), a recursive framework for long-form video-to-text summarization that extends the RLM paradigm to multimodal video reasoning. Importantly, the proposed approach is agnostic to the pretrained LLM used. RVLSM treats the input video as an external multimodal object and recursively summarizes semantically coherent video segments through VideoLLM sub-calls and external tools (see Figure~\ref{fig:framework}). Unlike static multimodal augmentation pipelines, RVLSM employs a heuristic routing mechanism to select the appropriate analysis configuration for each chunk based on previously generated summaries, enabling sequential context-aware evidence acquisition across the video timeline.
Specifically, RVLSM employs (1) \emph{Semantic Temporal Decomposition} (STD), which chunks videos according to scene transitions and additional contextual features to preserve temporal coherence, and (2) \emph{Adaptive Multimodal Tool Calling} (AMTC), which calls auxiliary external tools for object detection, action recognition, automatic speech recognition and optical character recognition 
through a programmable Root-LLM planner to ensure semantic richness of the summary.

\textit{RVLSM} addresses information compression and context saturation via the RLM paradigm, temporal coherence degradation using the STD component, and lack of semantic richness of the summary using the AMTC component.
Using \textit{RVLSM}, we address the following Research Questions: RQ1) Is RLM beneficial for LLM-based video-to-text summarization of long-form videos compared to non-recursive zero-shot baselines? (see the results in Table~\ref{tab:baseline_result}). RQ2) What are the effects of the main \textit{RVLSMs}'s components? (see the ablation study in Figure~\ref{fig:bertscore_comparison} and Table~\ref{tab:baseline_result}). RQ3) Are the \textit{RVLSMs}'s summaries more closely aligned with human-generated ones in terms of presence of different NLP entities? (see the results in Table~\ref{tab:ner_results}). 


\noindent The main paper \textbf{contributions} are summarized below:

\begin{itemize}
    \item We introduce the \textbf{Recursive Video Language Summarization Model} (RVLSM), a new recursive REPL-based framework for long-form video-to-text summarization agnostic to the pretrained LLMs used.
     \item On two video benchmarks,  
     we demonstrate that the \textbf{RVLSM consistently improves over zero-shot VideoLLM baselines} across multiple model settings and performance metrics , and is \textbf{competitive with a state-of-the-art fine-tuned model} for video-to-text summarization (e.g., SBERT F1 on MixtyVSum: RVLSM 0.64 vs. Baseline non-recursive 0.57 vs. V2Xum Baseline 0.44).
    
     \item We show \textbf{RVLSM's summaries align more closely with those detected in the ground-truth summaries} than those detected by baseline and state-of-the-art methods (e.g., F1 MicroAverage on MixtyTVSum: RVLSM 0.14 vs. Baseline non-recursive 0.07 vs. V2Xum Baseline 0.03), confirming the saliency of the event descriptions the model produces.
   
\end{itemize}


\section{Related Work}
\label{sec:rel}

\subsection{Video-to-text Summarization}
Video-to-text summarization has traditionally focused on extractive methods that select key frames or representative clips from long videos~\cite{NIPS2014_5d3b9e06,zhang2016video,LIN2022107618}. More recent approaches have introduced reinforcement learning for learning compact video representations~\cite{10.5555/3504035.3504964,WANG2024119888}. With the emergence of video LLMs, recent methods have shifted toward abstractive  video-to-text summarization, directly generating natural language descriptions from visual and multimodal inputs~\cite{mu2024realizing,cagliero-etal-2025-detecting,lee2025video,sc10.1609/aaai.v39i4.32374}. Despite strong performance on short videos, current VideoLLMs remain constrained by limited temporal context windows and typically rely on sparse frame sampling or fixed temporal compression strategies.

\subsection{Long-form Video Understanding}
Recent work has studied how to extend VideoLLMs capabilities to longer temporal horizons
through memory modules, hierarchical architectures, sparse attention mechanisms,
or temporal compression strategies~\cite{wu2024longvideobench,song2024moviechat,Wang_Chen_Yue_Wang_Qiao_Wang_Wang_2026}. Hierarchical summarization pipelines commonly split long videos into fixed-duration segments before independently summarizing each chunk~\cite{pennec-etal-2026-minimal}.
Although effective in reducing memory usage, fixed chunking may fragment semantically
coherent events across temporal boundaries. RVLSM, instead, performs scene-aware
temporal decomposition through dynamic scene-based chunking, enabling recursive
reasoning over semantically coherent video segments.

\subsection{Position of the Present Work} 
To the best of our knowledge, this is the first attempt to apply RLMs to long-context inputs in the video modality, targeting video-to-text summarization because its programmatic and recursive nature can overcome the aforementioned limitations. 

\section{Recursive Video Language Summarization Model}
\label{sec:method}

Given a base VideoLLM $M$ with limited temporal context capacity, a \textit{Recursive Video Language Summarization Model} (\textit{RVLSM}) is an inference-time scaffold around $M$ that treats the input video as part of an external environment while preserving the ability to densely process its temporal structure through recursive sub-calls. Given an input video

\[
V = \{f_1, f_2, \dots, f_T\},
\]

where $f_i$ denotes the $i_{th}$ video frame and its associated audio stream, RVLSM interacts with a persistent REPL environment $E$ to generate an abstractive summary $S$.
According to the RLM paradigm, instead of directly loading the entire video into the context window of a VideoLLM, RVLSM stores the video as an external object inside the REPL environment. The Root-LLM recursively decomposes the input video into semantically coherent temporal regions and delegates chunk-level summarization to 
sub-LLM calls.

Algorithm~\ref{alg:RVLSM} describes the overall execution process. Given a video $V$,
RVLSM initializes a REPL environment containing the input video together with functions
for scene-aware chunking, recursive VideoLLM querying, and multimodal tool execution.
The Root-LLM iteratively dispatches sub-calls to VideoLLM sub-models, collecting chunk-level summaries, and finally aggregates them into a single coherent abstractive summary.

\begin{algorithm}[t]
\caption{Recursive Video Language Summarization Model}
\small
\label{alg:RVLSM}
\begin{algorithmic}[1]
\Require Input video $V$, query $q$
\Ensure Final summary $S$

\State $E \gets \mathrm{InitREPL}(V)$
\State $E \gets \mathrm{AddFunctions}(E)$
\State $(c_1,\ldots,c_N) \gets \mathrm{ChunkVideoByScenes}(V)$
\State $\mathcal{S} \gets [\,]$

\For{$i = 1,\ldots,N$}
    \If{$i = 1$}
        \State $r_i \gets \mathrm{EnrichedVideoQuery(c_i, q)}$
    \Else
        \State $r_i \gets \mathrm{AdaptiveToolSelector}(s_{i-1})$
    \EndIf

    \State $s_i \gets r_i(c_i, q)$
    \State $\mathcal{S} \gets \mathcal{S} \,\Vert\, [s_i]$
\EndFor

\State $S \gets \mathrm{RootLLM}(\mathcal{S}, q)$
\State \Return $S$
\end{algorithmic}
\end{algorithm}

\noindent RVLSM introduces two key modifications tailored to long-video understanding: scene-aware \textit{Semantic Temporal Decomposition} (STD) and \textit{Adaptive Multimodal Tool Calling} (AMTC).

\subsection{Semantic Temporal Decomposition}

Unlike standard fixed-duration chunking, the STD module performs an adaptive scene-based segmentation of the video.
Given an input video $V$, the chunking module produces a sequence of chunks:

\[
V = \{c_1, c_2, \dots, c_N\}
\]

where each chunk $c_i$ corresponds to a scene-consistent temporal segment. A minimum chunk duration of 30 seconds is enforced, ensuring that no segment is too short to carry meaningful semantic content. All chunks are 
temporally disjoint, chronologically ordered, and exhaustive over the input video.


\subsection{REPL Environment and Recursive Video Queries}

RVLSM operates inside a persistent Python REPL environment containing the input video and a collection of executable functions. The environment exposes functions for retrieving video metadata, performing scene-aware semantic temporal decomposition, querying VideoLLMs, and extracting multimodal signals from video chunks. The Root-LLM interacts with the environment iteratively by generating executable Python code. Intermediate summaries and tool outputs are stored as variables inside the REPL, allowing subsequent recursive calls to access previously computed information.

For each chunk $c_i$, RVLSM invokes a sub-VLLM:
\[
s_i = \mathcal{M}_{sub}(c_i, q_i, k_i),
\]

\noindent where $\mathcal{M}_{sub}$ denotes the sub-VLLM, $q_i$ is the chunk-specific summarization query, and $k_i$ denotes optional multimodal information extracted
through external tools. Experiments focus on a single recursive level in which the Root-LLM delegates chunk-level summarization to VideoLLM sub-calls. The framework generalizes to deeper recursive hierarchies over temporal sub-events.

\subsection{Adaptive Multimodal Tool Calling} 
To complement recursive calls to VideoLLMs, RVLSM augments chunk-level analysis through external tools specialized for video understanding. 
The REPL environment provides access to auxiliary multimodal analysis functions that extract semantic signals from each video chunk.

Specifically, RVLSM can invoke object detection, action recognition, automatic speech recognition (ASR), and Optical Character Recognition (OCR). The extracted knowledge is injected into the VideoLLM sub-model prompt alongside the video chunk, providing structured perceptual grounding that enhances the quality of the generated chunk-level summary.


\begin{algorithm}[t]
\caption{AdaptiveToolSelector}
\small
\label{alg:router}
\begin{algorithmic}[1]
\Require Previous chunk summary $s_{i-1}$
\Ensure Video Query function $r_i$
\State $\mathcal{C} \gets \{\text{actions},\text{speech},\text{text},\text{objects}\}$ \Comment{signal categories}
\State $R_c \gets$ predefined keyword root set, for each $c \in \mathcal{C}$
\State $\mathrm{SingleModalityQuery} \gets \{$
    \State \quad $\text{actions} \to \mathrm{ActionQuery}\ \text{(action recognition)},$
    \State \quad $\text{speech} \to \mathrm{ASRQuery}\ \text{(automatic speech recognition)},$
    \State \quad $\text{text} \to \mathrm{OCRQuery}\ \text{(optical character recognition)},$
    \State \quad $\text{objects} \to \mathrm{ObjectQuery}\ \text{(object detection)}$
\State $\}$
\For{$c \in \mathcal{C}$}
    \State $\text{raw}[c] \gets |\{r \in R_c : r \subseteq s_{i-1}\}|$ \Comment{keyword matches}
    \State $\text{scores}[c] \gets \text{raw}[c] \,/\, |R_c|$ 
\EndFor
\State $\text{total} \gets \sum_{c \in \mathcal{C}} \text{scores}[c]$
\State $\text{top} \gets \arg\max_{c \in \mathcal{C}} \text{scores}[c]$
\If{$\text{total} = 0$ \textbf{ or } $\text{scores}[\text{top}] < 0.05$}
    \State \Return $\mathrm{VideoQuery}$ \Comment{no specific category }
\ElsIf{$\text{scores}[\text{top}] < 0.6 \cdot \text{total}$}
    \State \Return $\mathrm{EnrichedVideoQuery}$ \Comment{no dominant category}
\Else
    \State \Return $\mathrm{SingleModalityQuery}(\text{top})$ \Comment{dominant category }
\EndIf
\end{algorithmic}
\end{algorithm}
Unlike static multimodal augmentation pipelines, RVLSM performs heuristic adaptive tool selection. The first chunk is analyzed using full multimodal extraction in order to establish an initial semantic context. For subsequent chunks the Root-LLM selects the appropriate analysis configuration based on the previous chunk summary:

\[
r_i = \mathcal{R}(s_{i-1}),
\]

where $\mathcal{R}$ denotes the routing mechanism.


The procedure described in Algorithm~\ref{alg:router} focuses on the \textsc{AdaptiveToolSelector}. Each perceptual signal is associated with one of four categories (actions, speech,text, and objects) corresponding respectively to action recognition, automatic speech recognition, optical character recognition, and object detection. Every category is paired with a predefined set of representative keywords (its root set), against which the previous chunk summary $s_{i-1}$ is matched. A normalized keyword-matching score is then computed for each category by dividing the raw number of matches by the size of the corresponding keyword set, so that categories with larger keyword sets are not favored.
The selector defaults to VideoQuery when no signal is detected, to EnrichedVideoQuery when no category dominates, and to the corresponding SingleModalityQuery otherwise.

\subsection{Hierarchical Summary Aggregation}
After all chunks have been recursively summarized, the Root-LLM aggregates the local
summaries into a single global summary. Let

\[
\mathcal{S} = \{s_1, s_2, \dots, s_N\}
\]

be the ordered set of chunk summaries produced by the recursive VideoLLM calls.
The Root-LLM receives the complete sequence $\mathcal{S}$ and synthesizes a final
abstractive summary:

\[
S = \mathcal{M}_{root}(s_1, s_2, \dots, s_N).
\]
The Root-LLM is directed to maintain chronological order and consistency in the final summary.


\section{Evaluation Setup}
\label{sec:setting}

\paragraph{Hardware}
All experiments were run on a machine equipped with an Intel\textsuperscript{\textregistered} Core\textsuperscript{TM} 
i9-10980XE CPU, $1 \times$ NVIDIA\textsuperscript{\textregistered} 
A40 48GB GPU, $128$ GB of RAM running Ubuntu $22.04$ LTS. 
The overall cost is approximately 1,400 GPU hours, 
with several runs across various combinations of dataset, model, and setting. 
On average, each run consumes approximately $9.0$M input tokens 
and ${\sim}200$K output tokens, resulting in an average API cost 
of $\$5.58$ per run across all tested models and datasets.

\paragraph{Datasets} We evaluate RVLSM on two long video-to-text summarization benchmarks: VideoXum~\cite{10334011} and MixTyVSum~\cite{cagliero-etal-2025-detecting}. VideoXum focuses on holistic narrative summarization of open-domain videos, while MixTyVSum emphasizes dense event-centric summarization with temporal and spatial annotations. Following our long-video setting, we restrict VideoXum to videos longer than 230 seconds and we removed MSVD\cite{hendria2025msvdindonesianbenchmarkmultimodalvideotext} videos from MixTyVSum, because they are all shorter than 20 seconds. Additional details are provided in the Supplementary Document. 

\paragraph{Performance Metrics}
We compare the generated summaries with the ground truth using the F1-Score values of the following established metrics: ROUGE-1, ROUGE-2, ROUGE-L~\cite{lin-2004-rouge} and SBERT~\cite{reimers-gurevych-2019-sentence}, computed as the cosine similarity between the Sentence-BERT embeddings of the generated and reference summaries. We further report METEOR~\cite{banerjee-lavie-2005-meteor}, BLEU-4~\cite{papineni-etal-2002-bleu}, and CIDEr~\cite{oliveira-dos-santos-etal-2021-cider}, three metrics commonly adopted for video captioning and summarization evaluation. 
For the Named Entity Recognition analysis, we report Precision, Recall, and F1 score. More details in the Supplementary Document.  
Aggregated values are computed as the micro-average over TP, FP, and FN across entity types.

\paragraph{Models}
\label{sec:models}
Following the RLM paradigm~\cite{zhang2026recursivelanguagemodels}, our pipeline distinguishes between two types of models. The \text{Root-LLM} is a coding-oriented LLM acting as the main orchestrator: it decomposes the input, 
manages intermediate variables, and dispatches sub-calls within the REPL 
environment. The \text{sub-LLM} is, in our context, a multimodal Video LLM invoked recursively 
by the root-LLM to process individual video segments and return structured descriptions.
We experimented with two root-LLMs (GPT-5.4~\cite{singh2026openaigpt5card} and 
Qwen3-Coder-480B-A35B~\cite{qwenteam2025qwen3coder}) and four sub-LLMs (Molmo2-8B~\cite{clark2026molmo2openweightsdata}, Qwen3-VL-8B-Instruct, Qwen/Qwen3-VL-30B-Instruct~\cite{bai2025qwen3vltechnicalreport}, and Qwen/Qwen3-Omni-30B-A3B-Instruct~\cite{xu2025qwen3omnitechnicalreport}). 
Further details are given in the Supplementary Document. 

\paragraph{Auxiliary Modules}
For Named Entity Recognition we use SpaCy~\cite{Honnibal_spaCy_Industrial-strength_Natural_2020} (\textit{en\_core\_web\_sm} model).
For scene-aware STD we use PySceneDetect~\cite{castellano_pyscenedetect}.
Object-level semantic signals are extracted using YOLOv8~\cite{ultralytics_yolov8} pretrained on the COCO dataset, applied every 10 frames.
Action-level semantic evidence is obtained through MMAction2~\cite{mmaction2} using a TSN model (ResNet-50, Kinetics-400), retaining only predictions above a confidence threshold of $0.80$.
Automatic speech recognition is extracted using Whisper~\cite{radford2022robustspeechrecognitionlargescale} (\textit{medium} variant), and optical character recognition is detected using EasyOCR~\cite{easyocr}.

\section{Main Results and Discussion}
\label{sec:results}


\paragraph{Peformance comparison (RQ1): Is RLM beneficial for LLM-based video-to-text summarization of long-form videos compared to non-recursive zero-shot baselines?}

Table~\ref{tab:baseline_result} reports the main summarization results on MixTyVSum and VideoXum across the different sub-VLLM settings. 
Recursive video-to-text summarization outperforms the monolithic zero-shot baselines, demonstrating the benefit of STD for long videos. In fact, on MixTyVSum, the strongest performance is obtained by RVLSM with STD paired with Qwen3-Omni-30B, which achieves the highest ROUGE-1, ROUGE-2, and ROUGE-L scores; on VideoXum, RVLSM with Qwen3-VL-30B yields the best results overall, attaining both the highest ROUGE scores and the highest SBERT similarity. METEOR follows the same ordering as ROUGE across both benchmarks, corroborating this ranking.

\noindent The comparison with the fine-tuned V2Xum-LLaMA-7B~\cite{sc10.1609/aaai.v39i4.32374} is informative for assessing generalization. On VideoXum, on which it was optimized, the fine-tuned model retains an advantage on surface-overlap metrics, yet the best zero-shot RVLSM configuration still surpasses it on ROUGE-1, METEOR, and, most notably, on SBERT similarity (0.6380 vs.\ 0.5573), indicating that recursive summarization captures the semantics of the reference summaries even without task-specific training. On MixTyVSum the picture reverses: the fine-tuned model falls below the zero-shot sub-VLLMs on nearly all metrics, whereas RVLSM improves over both baselines by a wide margin. This suggests that a model tuned to a single summarization distribution transfers poorly across benchmarks, while the recursive approach generalizes robustly.

\begin{figure}[th!]
    \centering
    \includegraphics[width=0.5\textwidth]{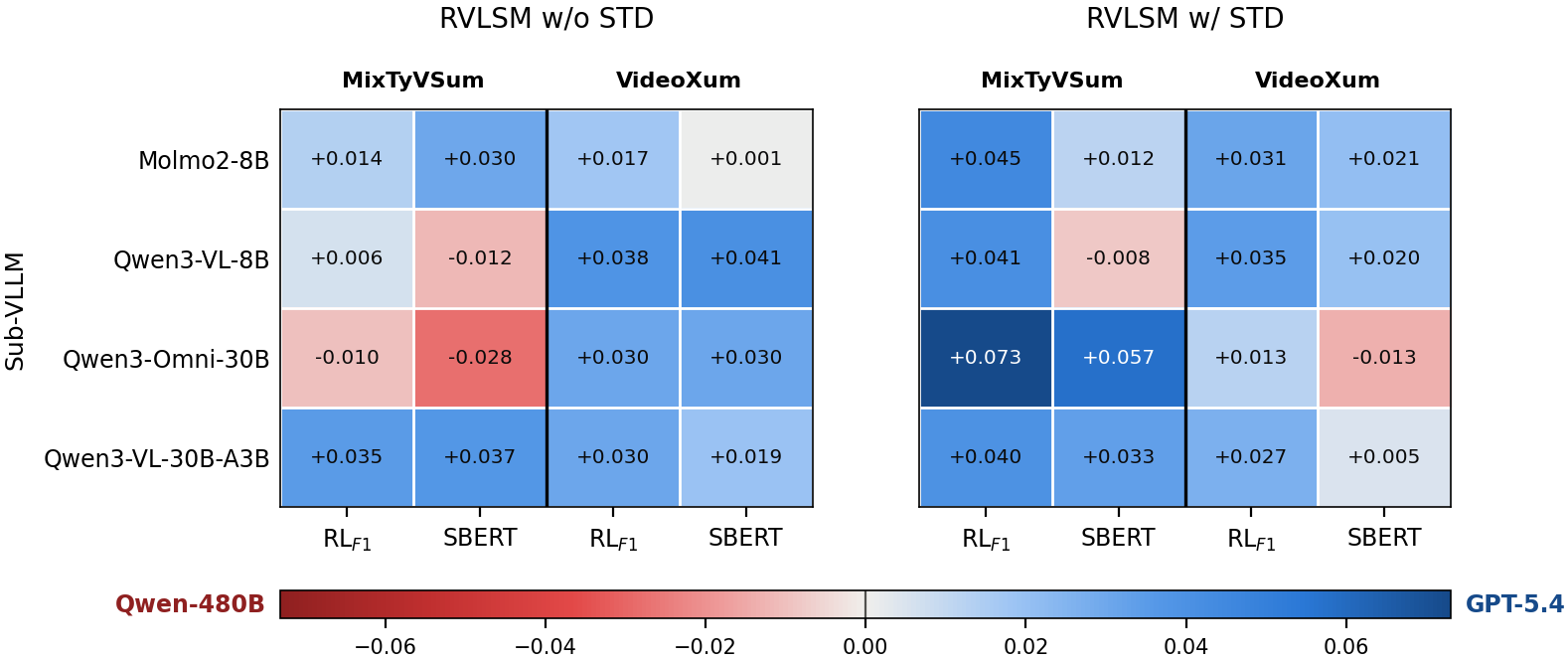}
  \caption{Proprietary and open-source Root-LLM: performance gap between GPT-5.4 and Qwen-480B across datasets and settings (w/ and w/o STD).}
    \label{fig:root_comparison}
\end{figure}

\paragraph{Choice of the Root-LLM: Proprietary vs. Opensource}
The choice of Root-LLM markedly affects recursive summarization quality. Figure~\ref{fig:root_comparison} shows, for each combination of sub-VLLM and metric, the differences in performance scores between using GPT-5.4 and Qwen-480B as Root-LLM, 
under both fixed-duration (w/o STD) and scene-aware (w/ STD) decomposition. Results show that the peformance is roughly comparable, underscoring the practical advantages of using opensource tools.

\begin{table*}[th!]
  \small
  \centering
\begin{tabular}{ll ccccccc}
\toprule
\multirow{1}{*}{\text{Sub-VLLM}} & \multirow{1}{*}{\text{Method}} & \text{R1$_{F1}$ $\uparrow$} & \text{R2$_{F1}$ $\uparrow$} & \text{RL$_{F1}$ $\uparrow$} & \text{SBERT $\uparrow$} & \text{METEOR $\uparrow$} & \text{BLEU-4 $\uparrow$} & \text{CIDEr $\uparrow$} \\ \midrule

\multicolumn{9}{c}{\textbf{MixTyVSum  dataset}} \\ \midrule

\multicolumn{9}{c}{\textit{Fine-Tuned Baseline}} \\
\multicolumn{2}{l}{V2Xum-LLaMA-7B (Fine-Tuned)} & 0.2115 & 0.0392 & 0.1603 & 0.4463 & \textbf{0.1092} & 0.0043 & 0.0152 \\ \midrule
\multicolumn{9}{c}{\textit{Zero-Shot Baselines}} \\
\multicolumn{2}{l}{Molmo2-8B}        & 0.3212 & 0.0694 & 0.2018 & 0.5445 & 0.1416 & 0.0124 & 0.0260 \\
\multicolumn{2}{l}{Qwen3-VL-8B}      & 0.2669 & 0.0504 & 0.1797 & 0.5664 & 0.1245 & 0.0169 & 0.0474 \\
\multicolumn{2}{l}{Qwen3-Omni-30B}   & 0.2946 & 0.0682 & 0.1947 & 0.5788 & 0.1624 & 0.0137 & 0.0435 \\
\multicolumn{2}{l}{Qwen3-VL-30B} & 0.3132 & 0.0736 & 0.2057 & 0.5794 & 0.1538 & 0.0146 & 0.0459 \\ \midrule
\multicolumn{9}{c}{\textit{RVLSM} (ours)} \\
\multirow{3}{*}{Qwen3-Omni-30B} & w/ STD & \textbf{0.3606} & \textbf{0.0920} & \textbf{0.2298} & 0.5859 & 0.1527 & \textbf{0.0149} & 0.0542 \\
 & +MTC  & 0.3233 & 0.0719 & 0.2083 & 0.5778 & 0.1737 & 0.0130 & 0.0536 \\
 & +AMTC & 0.3068 & 0.0643 & 0.1924 & 0.6063 & 0.1652 & 0.0120 & \textbf{0.0570} \\ 
\multirow{3}{*}{Qwen3-VL-30B} & w/ STD & 0.3234 & 0.0753 & 0.2113 & 0.5958 & 0.1547 & 0.0121 & 0.0513 \\
 & +MTC  & 0.3303 & 0.0803 & 0.2142 & 0.6149 & 0.1724 & 0.0140 & 0.0568 \\
 & +AMTC & 0.3192 & 0.0683 & 0.2051 & \textbf{0.6401} & \textbf{0.1801} & 0.0146 & 0.0531 \\ \midrule

\multicolumn{9}{c}{\textbf{VideoXum dataset}} \\ \midrule

\multicolumn{9}{c}{\textit{Fine-Tuned Baseline}} \\
\multicolumn{2}{l}{V2Xum-LLaMA-7B (F.T.)} & 0.3228 & 0.1016 & 0.2475 & 0.5573 & 0.1826 & \textbf{0.0292} & \textbf{0.1386} \\ \midrule
\multicolumn{9}{c}{\textit{Zero-Shot Baselines}} \\
\multicolumn{2}{l}{Molmo2-8B}        & 0.2860 & 0.0651 & 0.1843 & 0.5975 & 0.1920 & 0.0142 & 0.0208 \\
\multicolumn{2}{l}{Qwen3-VL-8B}      & 0.2761 & 0.0428 & 0.1840 & 0.5799 & 0.1558 & 0.0122 & 0.0478 \\
\multicolumn{2}{l}{Qwen3-Omni-30B}   & 0.3041 & 0.0606 & 0.2001 & 0.5949 & 0.1913 & 0.0146 & 0.0344 \\
\multicolumn{2}{l}{Qwen3-VL-30B} & 0.3230 & 0.0715 & 0.2154 & 0.6053 & 0.1958 & \textbf{0.0224} & 0.0544 \\ \midrule
\multicolumn{9}{c}{\textit{RVLSM} (ours)} \\
\multirow{3}{*}{Qwen3-Omni-30B} & w/ STD & 0.3035 & \textbf{0.0584} & 0.2011 & 0.5911 & 0.1806 & 0.0143 & 0.0681 \\
 & +MTC  & 0.2929 & 0.0497 & 0.1962 & 0.5840 & 0.1552 & 0.0108 & 0.0506 \\
 & +AMTC & 0.2815 & 0.0449 & 0.1929 & 0.6010 & 0.1912 & 0.0121 & 0.0356 \\ 
\multirow{3}{*}{Qwen3-VL-30B} & w/ STD & \textbf{0.3326} & \textbf{0.0745} & \textbf{0.2257} & \textbf{0.6380} & \textbf{0.2023} & 0.0176 & 0.0739 \\
 & +MTC  & 0.3024 & 0.0571 & 0.2077 & 0.6094 & 0.1589 & 0.0124 & \textbf{0.0714} \\
 & +AMTC & 0.3100 & 0.0595 & 0.2031 & 0.6134 & 0.1873 & 0.0132 & 0.0532 \\ \bottomrule
\end{tabular}
  \caption{Performance comparison between RVLSM (ours), the traditional (non-recursive) VideoLLM baselines, and the recently proposed V2Xum model~\cite{sc10.1609/aaai.v39i4.32374} fine-tuned on VideoXum. 
  The \textit{RVLSM} block reports the recursive pipeline with scene-aware STD, and its combination with multimodal tool calling (STD+MTC) and adaptive multimodal tool calling (STD+AMTC), across two representative Sub-VLLMs.
  }
\label{tab:baseline_result}
\end{table*}

\begin{figure}[t!]
    \centering

    \begin{subfigure}[t]{0.43\columnwidth}
        \centering
        \includegraphics[width=0.65\linewidth]{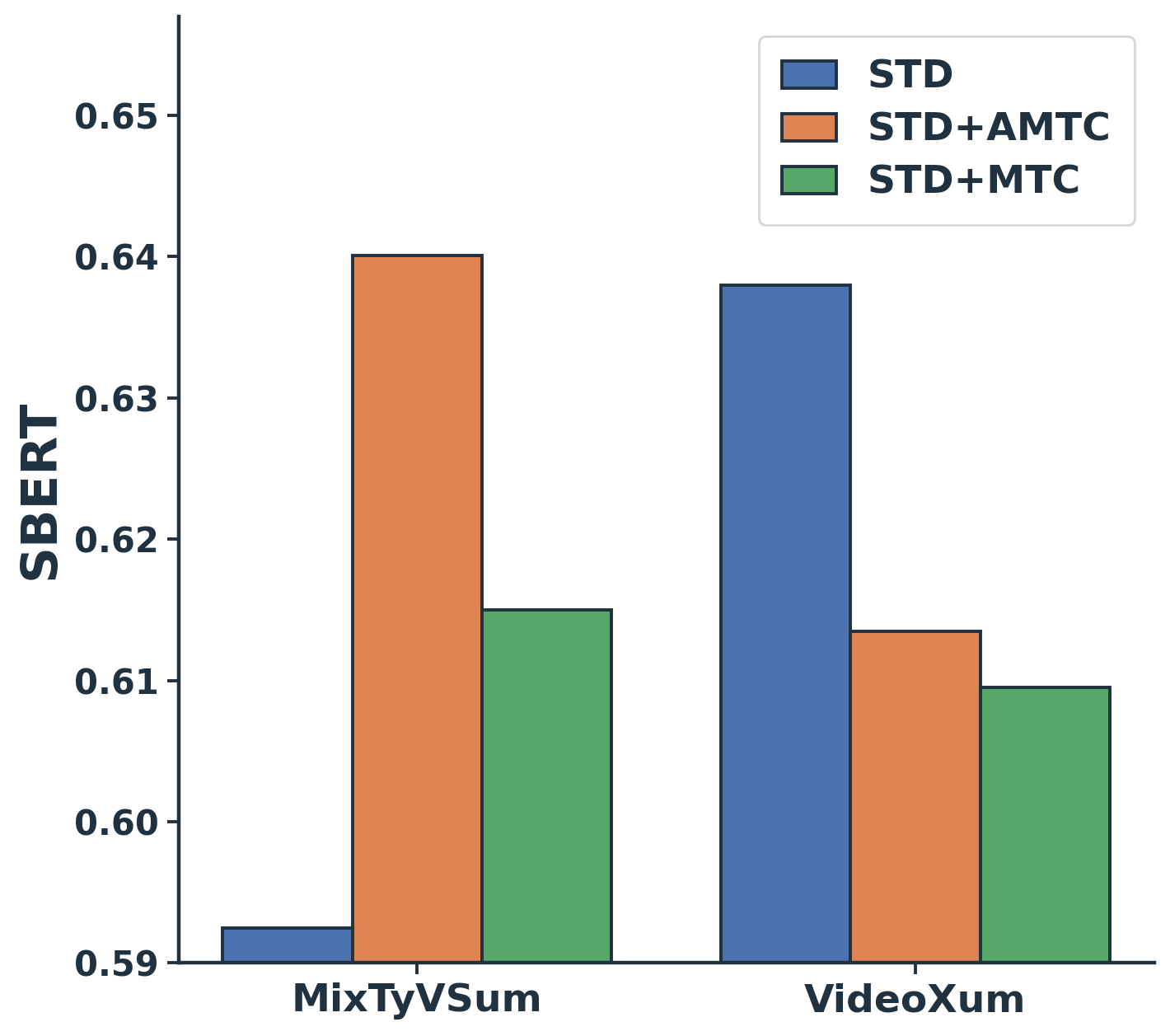}
        \caption{}
        \label{fig:MixVsVid}
    \end{subfigure}
    \hfill
    \begin{subfigure}[t]{0.56\columnwidth}
        \centering
        \includegraphics[width=0.95\linewidth]{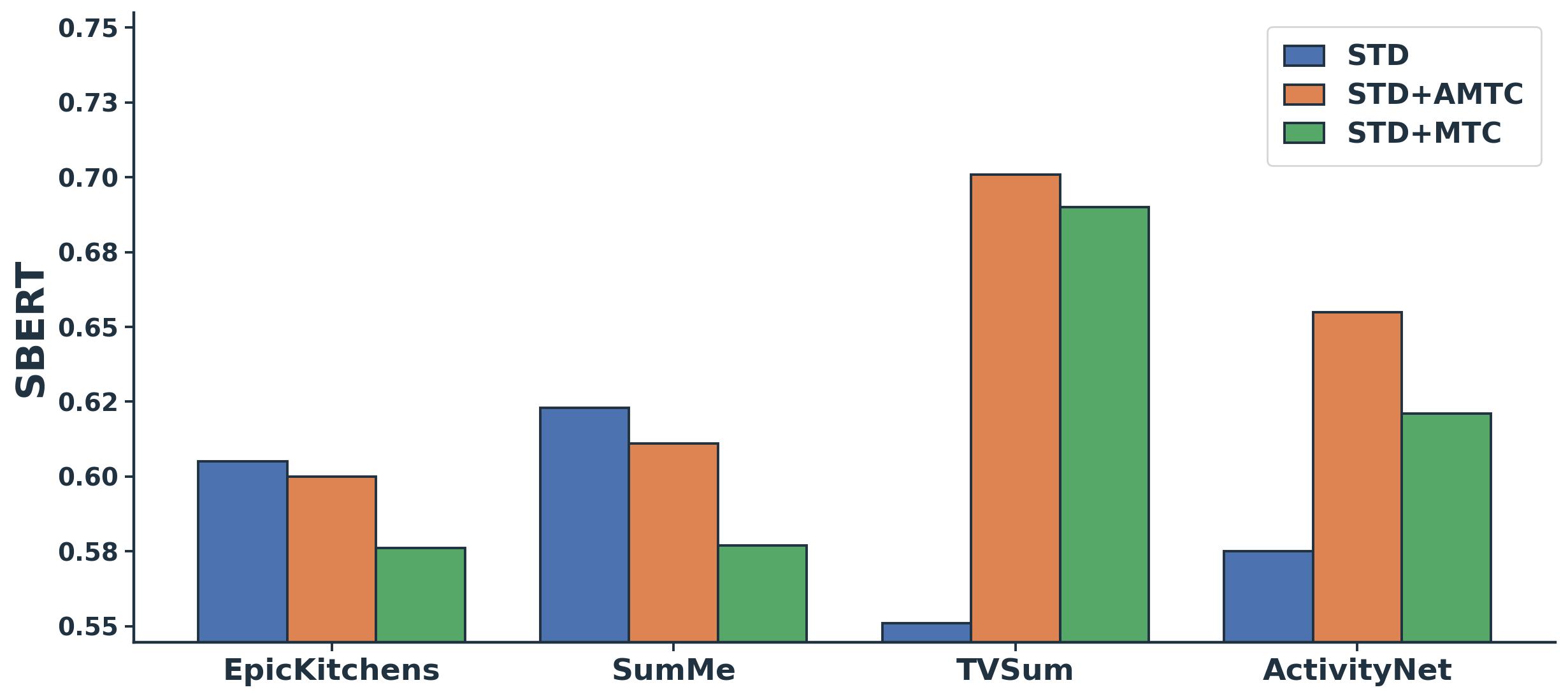}
        \caption{}
        \label{fig:MixTy}
    \end{subfigure}

   \caption{SBERT comparison on MixTyVSum and VideoXum (a) and across the MixTyVSum subsets (b).}
    \label{fig:bertscore_comparison}
\end{figure}

\paragraph{Ablation study (RQ2): what are
the effects of the main RVLSMs’s components? }
\label{sec:ablation}
We examine the role of Adaptive Multimodal Tool Calling (AMTC) by comparing three scenarios: RVLSM+STD, which employs scene-aware decomposition without multimodal tool integration; RVLSM+MTC, which uses multimodal tools but not adaptively; and RVLSM+AMTC, which employs multimodal tools adaptively. Table~\ref{tab:baseline_result} reports the performance of the two sub-VLLMs under these strategies, and Figure~\ref{fig:MixVsVid} summarizes the corresponding SBERT behavior on the two benchmarks.

\noindent A clear divergence arises between the benchmarks due to differing annotation styles. On MixTyVSum, which features action-oriented and semantically dense summaries, adaptive injection enhances semantic quality (SBERT and METEOR). 
However, this can lead to a decline in ROUGE scores, particularly for Qwen3-Omni-30B, which performs best with STD alone, indicating a trade-off between semantic detail and lexical overlap. Conversely, on VideoXum, where summaries are more generic, injection is largely ineffective, with RVLSM+STD achieving the best scores across most metrics for both sub-VLLMs.

Additionally, Figure~\ref{fig:MixTy} decomposes SBERT by source dataset and shows that the benefit of injection depends on how well STD alone already captures the video content. On structured, procedural subsets such as EpicKitchens and SumMe, STD is already strong and injection yields no gain or a marginal drop; on more diverse subsets such as TVSum and ActivityNet, adaptive injection provides substantial improvements (TVSum: STD=0.551 to AMTC=0.701; ActivityNet: STD=0.575 to AMTC=0.655). Across all four subsets, RVLSM+AMTC consistently outperforms RVLSM+MTC, and its advantage is largest precisely in the structured domains where indiscriminate injection is most harmful (EpicKitchens: MTC=0.577 to AMTC=0.600; SumMe: MTC=0.577 to AMTC=0.611). This confirms the benefit of adaptive tool selection over unconditional invocation: applying every multimodal analysis regardless of context introduces noise that offsets, and in structured domains, outweighs the benefit of the injected signal.

\paragraph{Semantic alignment (RQ3): are RVLSMs’s summaries closely aligned with human-generated ones?}
\label{sec:entity}

We empirically verify whether the \textit{RVLSM}'s summaries are more closely aligned with human expectations 
than baseline methods in terms of NLP entities retrieved in the ground-truth summary.
The core idea is to verify the semantic alignment in terms of matching with semantically rich concepts.
Table~\ref{tab:ner_results} reports the Precision, Recall, and F1 MicroAverages separately for each entity type. 

On MixTyVSum, RVLSM attains the best aggregate scores (F1=0.170 to 0.081 for the VideoLLM baseline) and leads on most entity types, including \textsc{work\_of\_art}, \textsc{date}, and \textsc{person}, where the baselines recover no entity. Overall, RVLSM recovers salient entities that surface-overlap metrics tend to overlook. We also evaluate extracted named entities on  V2Xum-generated summaries. However, only a few \textsc{cardinal} are detected, leading to all zeros matric values for all other entities. 
More results are reported in the Supplementary Document.

\begin{table}[t!]
  \centering
  \small
  \setlength{\tabcolsep}{2.5pt}
\begin{tabular}{l ccc ccc}
\toprule
\multirow{2}{*}{\text{Entity type}}
  & \multicolumn{3}{c}{\textit{RVLSM (ours)}}
  & \multicolumn{3}{c}{\textit{Baseline VideoLLM}} \\
\cmidrule(lr){2-4} \cmidrule(lr){5-7}
  & P $\uparrow$ & R $\uparrow$ & F1 $\uparrow$ & P $\uparrow$ & R $\uparrow$ & F1 $\uparrow$ \\
\midrule
\texttt{CARDINAL}      & 0.393 & \textbf{0.204} & \textbf{0.268} & 0.125 & 0.024 & 0.041 \\
\texttt{GPE}           & 0.188 & \textbf{0.300} & 0.231          & \textbf{0.333} & 0.250 & \textbf{0.286} \\
\texttt{WORK\_OF\_ART} & \textbf{0.308} & \textbf{0.190} & \textbf{0.235} & 0.000 & 0.000 & 0.000 \\
\texttt{DATE}          & \textbf{0.222} & \textbf{0.182} & \textbf{0.200} & 0.143 & 0.100 & 0.118 \\
\texttt{PERSON}        & \textbf{0.128} & \textbf{0.312} & \textbf{0.182} & 0.000 & 0.000 & 0.000 \\
\texttt{EVENT}         & 0.000 & 0.000 & 0.000 & 0.000 & 0.000 & 0.000 \\
\texttt{ORG}           & \textbf{0.083} & \textbf{0.235} & \textbf{0.123} & 0.000 & 0.000 & 0.000 \\
\texttt{NORP}          & 0.000 & 0.000 & 0.000 & \textbf{0.200} & \textbf{0.500} & \textbf{0.286} \\
\texttt{ORDINAL}       & 0.167 & 0.027 & 0.047 & \textbf{1.000} & \textbf{0.031} & \textbf{0.061} \\
\texttt{FAC}           & 0.000 & 0.000 & 0.000 & \textbf{0.500} & \textbf{1.000} & \textbf{0.667} \\
\midrule
\textit{Mean}          & \textbf{0.170} & \textbf{0.170} & \textbf{0.170} & 0.136 & 0.058 & 0.081 \\
\bottomrule
\end{tabular}
  \caption{Precision, Recall, and F1 of the entities detected on MixTyVSum by RVLSM (GPT-5.4 root, Qwen3-VL-30B sub-model) and the VideoLLM baseline (same sub-model). 
  Best value per entity type and metric in bold. }
  \label{tab:ner_results}
\end{table}

\subsection{LLM-as-a-Judge}
\label{sec:llm-as-a-judge}
To complement automatic metrics, an LLM-as-a-Judge A/B test~\cite{10.5555/3666122.3668142} is conducted using Gemini 3.1 Pro (\texttt{gemini-3.1-pro-preview}) across all three configurations \texttt{Qwen3-VL-30B} compared with the baseline. Results in Table~\ref{tab:llm_judge} confirm the trends 
observed through automatic metrics, with the use of STD and the AMTC achieving the best overall performance. 

\begin{table}[t!]
\centering
\small
\begin{tabular}{lcc}
\hline
\textbf{Selected Methods} & \textbf{Wins} & \textbf{Win \%} \\
\hline
RVLSM\textsubscript{w/ STD}     & 106 & 34.2\% \\
RVLSM\textsubscript{AMTC}      & 91 & 29.2\% \\
RVLSM\textsubscript{w/o STD}       & 69 & 22.2\% \\
Baseline                                & 45 &  14.4\% \\
\hline
\end{tabular}
\caption{LLM-as-a-Judge evaluation on the VideoXum and MixTyVSum datasets, using Qwen3-VL-30B as sub-VLLM and GPT-5.4 as Root-LLM.}
\label{tab:llm_judge}
\end{table}
\begin{figure}[h]
    \centering
    \includegraphics[width=0.4\textwidth]{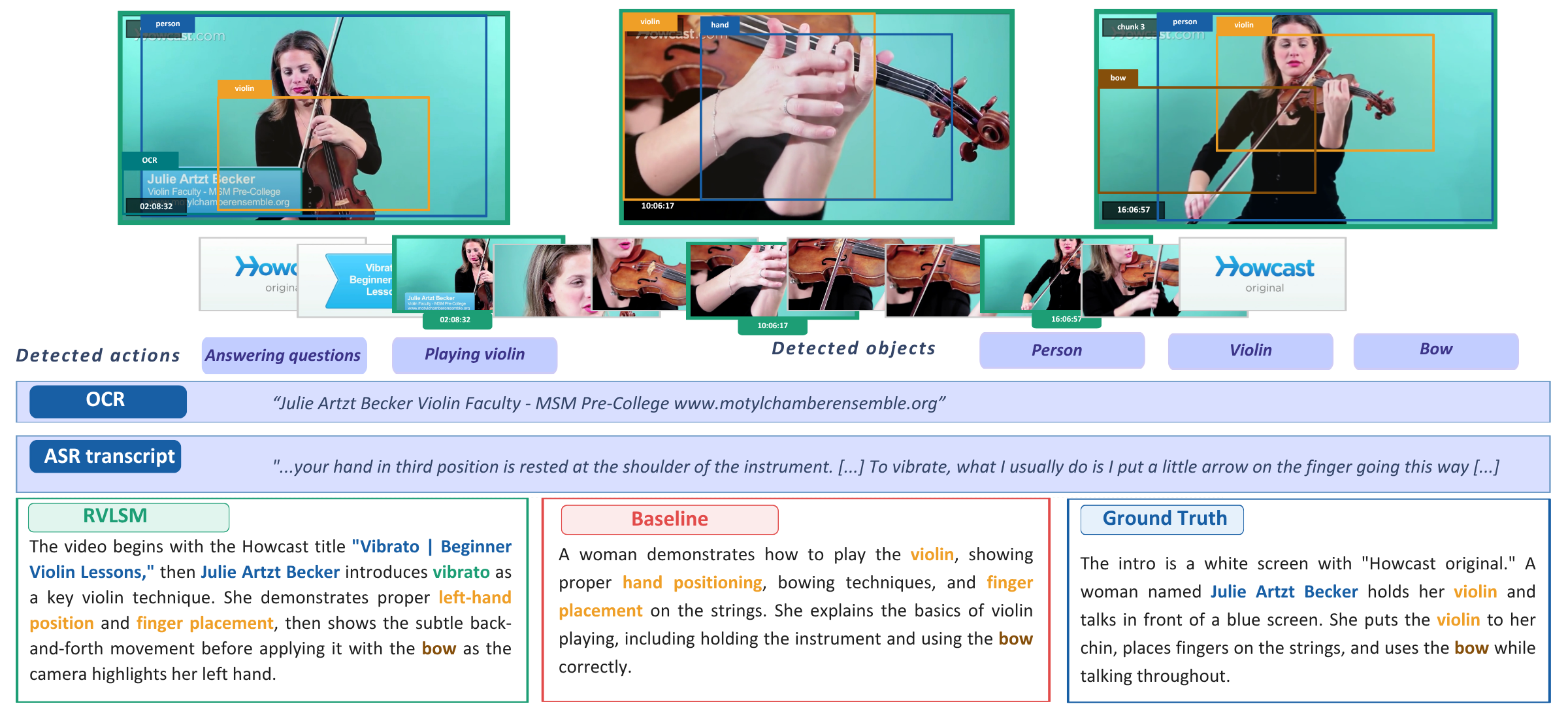}
    \caption{A qualitative example from VideoXum showing the summaries generated by RVLSM w/STD + AMTC, compared to the zero-shot baseline and the ground truth}
    \label{fig:tvsum_example}
\end{figure}
\subsection{Qualitative Analysis}
\begin{figure}[h]
    \centering
    \includegraphics[width=0.4\textwidth]{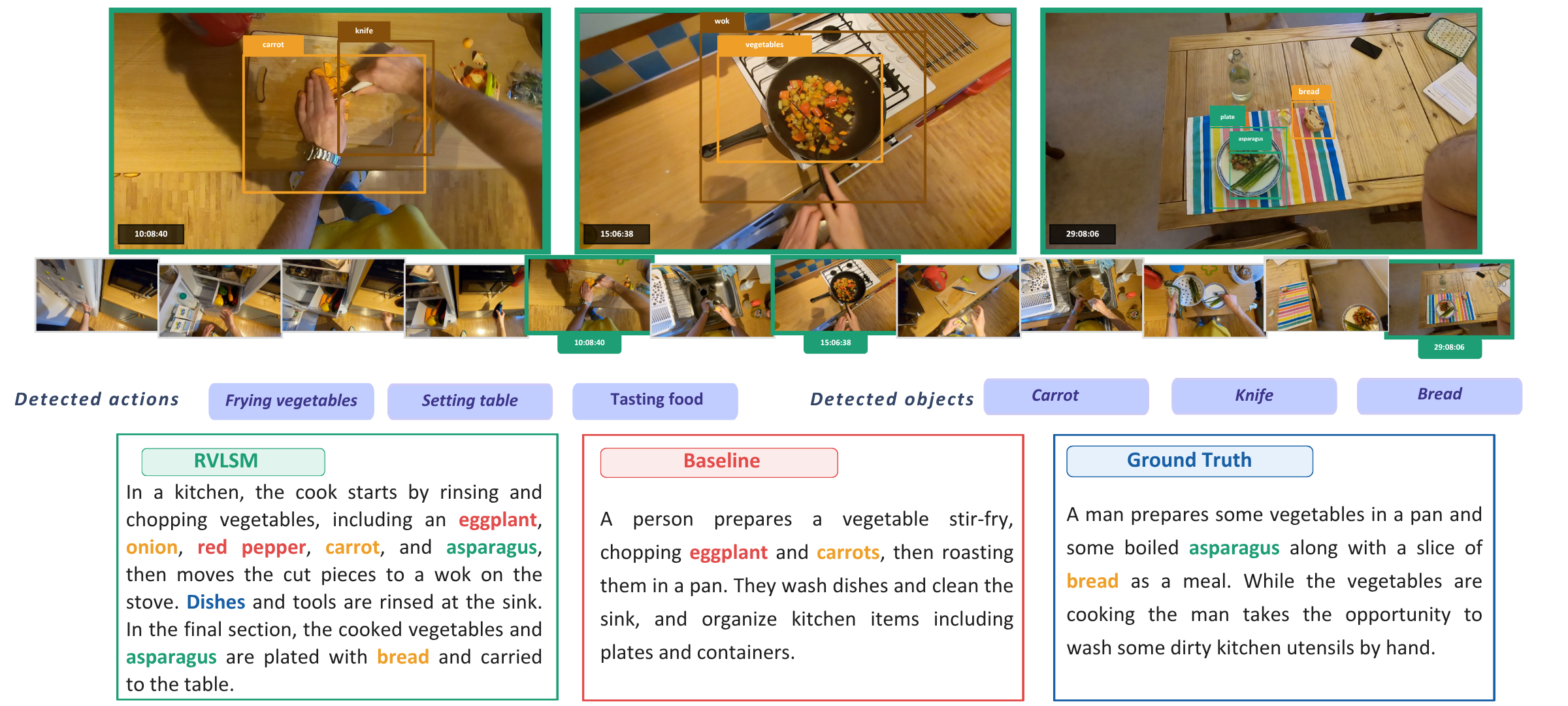}
    \caption{ A qualitative example from MixTyVSum showing the summaries generated by RVLSM w/ STD + AMTC, compared to the zero-shot baseline and the ground truth.}
    \label{fig:epic_example}
\end{figure}
This section presents two representative examples highlighting scenarios in which the RVLSM approach shows clear advantages over the baseline. Figure~\ref{fig:tvsum_example} reports the result generated from a sample video of the VideoXum dataset, where the RVLSM summary exploits OCR injection to accurately identify the on-screen information and the ASR transcript to detail the specific technique demonstrated. In contrast, the baseline produces a generic description that mentions neither the instructor's identity nor the specific content, confirming that without recursive decomposition relevant information dispersed over time tends to be lost. A second case is shown in Figure~\ref{fig:epic_example}, a summary generated from EpicKitchens as part of the MixTyVSum annotations. Here the baseline captures only a subset of ingredients and omits secondary activities such as dishwashing and table setting, whereas the RVLSM, leveraging its recursive structure, reconstructs the full sequence of events in close alignment with the ground truth. This example illustrates how the framework scales with video duration, yielding greater improvements as content becomes longer and more complex.
\subsection{Statistical Significance Tests}
\label{sec:test}
To rank the summarizers on each benchmark, we apply the Borda Count re-ranking strategy~\cite{borda}, a group consensus function that merges the per-video rankings (231 for VideoXum, 60 for MixTyVSum) into a single one by assigning decreasing integer scores to each ranking and summing them. The RVLSM approaches w/ STD dominate on both benchmarks: on MixTyVSum, Qwen3-Omni-30B and Qwen3-VL-30B rank first and second (242 and 221) ahead of the best w/o STD model (205), and Qwen3-VL-30B leads on VideoXum.

\section{Conclusions and future work}
\label{sec:concl}
We introduced the recursive Video Language Summarization Model (RVLSM), which extends the RLM paradigm to long-video abstractive summarization through Semantic Temporal Decomposition (STD) and Adaptive Multimodal Tool Calling (AMTC). Across multiple VideoLLMs and benchmarks, RVLSM yields consistent improvements that grow with video duration, while our ablation shows that the value of knowledge injection depends on the informativeness of the visual signal, motivating adaptive rather than indiscriminate augmentation. Future work will explore deeper recursion, aggregating summaries over multiple hierarchical levels, and Deep Reinforcement Learning to mitigate specific limitations of the RLM approach.

\section{Limitations}
\label{sec:limitations}
Despite promising results, our approach has several limitations that point to directions for future work.
\paragraph{Routing Mechanism.} The current routing relies on heuristic keyword matching; more principled alternatives, such as learned or reinforcement-learning-based routing optimized for summarization quality, are a promising direction.
\paragraph{Dataset Availability.} The benefits of RVLSM would likely be more pronounced on longer videos, yet benchmarks with very long videos and high-quality reference summaries remain scarce due to the annotation effort required.
\paragraph{Chunking Quality.} Visual scene detection may yield semantically incoherent segments when transitions are subtle, as in egocentric videos; multimodal boundary detection combining visual, audio, and action signals could mitigate this.
\paragraph{Annotation Sensitivity.} Evaluation depends on the annotation style of the reference summaries: generic and event-driven ones respond very differently to knowledge injection, suggesting that standard metrics may not assess model quality in an annotation-agnostic way.
\section*{Ethics, risks, and Generative AI Statement}
The video data and annotations used in this work were obtained from third-party sources and do not introduce ethical concerns, such as unethical or offensive content, related to the input data. Due to the non-deterministic nature of the generative process, we cannot ensure that the summarized outputs are free of hallucination. However, the output summaries have been reviewed by at least two PhD-level researchers prior to the final evaluation. Generative AI was used as a research engine and its outputs were carefully reviewed by the authors; minor AI support also includes writing suggestions such as grammar corrections. The project code is publicly available under the CC BY-NC license. Part of the project builds upon the original RLM architecture~\cite{zhang2026recursivelanguagemodels}; credits to the owner, changes, and specific settings are reported in the code repository.

\bibliography{aaai2027}


\end{document}